\documentclass[letterpaper, 10 pt, conference]{class/ieeeconf}  

\IEEEoverridecommandlockouts                              
\usepackage{xcolor}
\usepackage{siunitx}  
\usepackage{graphicx}
 \graphicspath{./figures/robot_demonstration}

\usepackage{amsmath}
\usepackage{amssymb}
\usepackage{latexsym}
\usepackage{url}
\usepackage{cite}
\usepackage{relsize}
\usepackage{multirow}
\usepackage{afterpage}
\usepackage{ifthen}
\usepackage{graphicx}
\usepackage{algpseudocode,algorithm,algorithmicx}
\usepackage{subfigure}
\usepackage{epstopdf}
\usepackage{stackengine}
\usepackage{textcomp} 

\usepackage{gensymb}
\usepackage{booktabs}
\usepackage{makecell}
\usepackage{hhline}
\usepackage{courier}
\usepackage{lipsum}
\usepackage{bbm}
\usepackage{bm}
\usepackage{pifont}

\newcommand{\xmark}{\ding{55}}%

\usepackage{xcolor} 
            
\makeatletter
\let\NAT@parse\undefined
\makeatother
\usepackage[bookmarks=false, linkcolor=blue, urlcolor=blue, citecolor=blue]{hyperref} 
\usepackage{etoolbox}
\makeatletter
\patchcmd{\@makecaption}
  {\scshape}
  {}
  {}
  {}
\makeatother
\hypersetup{
    colorlinks=true,
    linkcolor=red,
    filecolor=magenta,      
    urlcolor=blue,
    pdfstartview={FitH},
    citecolor =blue
    }

\DeclareRobustCommand{\vec}[1]{ 				
	\ifthenelse{\equal{#1}{\omega} \OR \equal{#1}{\varphi} \OR \equal{#1}{\alpha} \OR \equal{#1}{\beta} \OR \equal{#1}{\chi} \OR \equal{#1}{\delta} \OR \equal{#1}{\varepsilon} \OR \equal{#1}{\phi} \OR \equal{#1}{\epsilon} \OR \equal{#1}{\gamma} \OR \equal{#1}{\eta} \OR \equal{#1}{\iota} \OR \equal{#1}{\kappa} \OR \equal{#1}{\lambda} \OR \equal{#1}{\mu} \OR \equal{#1}{\nu} \OR \equal{#1}{\pi} \OR \equal{#1}{\theta} \OR \equal{#1}{\vartheta} \OR \equal{#1}{\rho} \OR \equal{#1}{\sigma} \OR \equal{#1}{\varsigma} \OR \equal{#1}{\tau} \OR \equal{#1}{\upsilon} \OR \equal{#1}{\xi} \OR \equal{#1}{\psi} \OR \equal{#1}{\zeta}}{
		\boldsymbol{#1}
	}{
		\mathbf{#1}
	}
}

\providecommand{\Comma}{\text{~,\xspace}}
\newcommand{\transpose}[1]{#1^\mathrm{T}}

\usepackage{xspace}
\DeclareMathOperator*{\argmax}{arg\,max}

 \graphicspath{{./figures/}}

\usepackage{tikz}
\usepackage{pgfplots}
\usepackage{pgfplotstable}
\newcommand\corr[1]{\textcolor{black}{#1}}  

\title{\LARGE \bf Towards Autonomous Wood-Log Grasping with a Forestry Crane: Simulator and Benchmarking}

\author{M.N. Vu$^{1,2}$, A. Wachter$^{1}$, G. Ebmer$^{1}$, M. Ecker$^{1,2}$, T. Gl{\"u}ck$^{2}$, A. Nguyen$^{3}$, \\W. Kemmetm{\"u}ller$^1$, and A. Kugi$^{1,2}$
    \thanks{$^{1}$Automation \& Control Institute (ACIN), TU Wien, 1040 Vienna, Austria {\tt\small vu@acin.tuwien.ac.at}}%
    \thanks{$^{2}$AIT Austrian Institute of Technology GmbH, 1210 Vienna, Austria}%
    \thanks{$^{3}$Department of Computer Science, University of Liverpool}
    }
\begin{document}

\newtheorem{problem}{Problem}
\newtheorem{lemma}{Lemma}
\newtheorem{theorem}[lemma]{Theorem}
\newtheorem{claim}{Claim}
\newtheorem{corollary}[lemma]{Corollary}
\newtheorem{definition}[lemma]{Definition}
\newtheorem{proposition}[lemma]{Proposition}
\newtheorem{remark}[lemma]{Remark}
\newenvironment{LabeledProof}[1]{\noindent{\it Proof of #1: }}{\qed}

\def\beq#1\eeq{\begin{equation}#1\end{equation}}
\def\bea#1\eea{\begin{align}#1\end{align}}
\def\beg#1\eeg{\begin{gather}#1\end{gather}}
\def\beqs#1\eeqs{\begin{equation*}#1\end{equation*}}
\def\beas#1\eeas{\begin{align*}#1\end{align*}}
\def\begs#1\eegs{\begin{gather*}#1\end{gather*}}

\newcommand{\poly}{\mathrm{poly}}
\newcommand{\eps}{\epsilon}
\newcommand{\e}{\epsilon}
\newcommand{\polylog}{\mathrm{polylog}}
\newcommand{\rob}[1]{\left( #1 \right)} 
\newcommand{\sqb}[1]{\left[ #1 \right]} 
\newcommand{\cub}[1]{\left\{ #1 \right\} } 
\newcommand{\rb}[1]{\left( #1 \right)} 
\newcommand{\abs}[1]{\left| #1 \right|} 
\newcommand{\zo}{\{0, 1\}}
\newcommand{\zonzo}{\zo^n \to \zo}
\newcommand{\zokzo}{\zo^k \to \zo}
\newcommand{\zot}{\{0,1,2\}}
\newcommand{\en}[1]{\marginpar{\textbf{#1}}}
\newcommand{\efn}[1]{\footnote{\textbf{#1}}}
\newcommand{\vecbm}[1]{\boldmath{#1}} 
\newcommand{\uvec}[1]{\hat{\vec{#1}}}
\newcommand{\thv}{\vecbm{\theta}}
\newcommand{\junk}[1]{}
\newcommand{\var}{\mathop{\mathrm{var}}}
\newcommand{\rank}{\mathop{\mathrm{rank}}}
\newcommand{\diag}{\mathop{\mathrm{diag}}}
\newcommand{\tr}{\mathop{\mathrm{tr}}}
\newcommand{\acos}{\mathop{\mathrm{acos}}}
\newcommand{\atantwo}{\mathop{\mathrm{atan2}}}
\newcommand{\SVD}{\mathop{\mathrm{SVD}}}
\newcommand{\quadf}{\mathop{\mathrm{q}}}
\newcommand{\linterp}{\mathop{\mathrm{l}}}
\newcommand{\sgn}{\mathop{\mathrm{sign}}}
\newcommand{\sym}{\mathop{\mathrm{sym}}}
\newcommand{\avg}{\mathop{\mathrm{avg}}}
\newcommand{\mean}{\mathop{\mathrm{mean}}}
\newcommand{\erf}{\mathop{\mathrm{erf}}}
\newcommand{\grad}{\nabla}
\newcommand{\R}{\mathbb{R}}
\newcommand{\defeq}{\triangleq}
\newcommand{\dims}[2]{[#1\!\times\!#2]}
\newcommand{\sdims}[2]{\mathsmaller{#1\!\times\!#2}}
\newcommand{\udims}[3]{#1}
\newcommand{\udimst}[4]{#1}
\newcommand{\com}[1]{\rhd\text{\emph{#1}}}
\newcommand{\ind}{\hspace{1em}}
\newcommand{\argmin}[1]{\underset{#1}{\operatorname{argmin}}}
\newcommand{\floor}[1]{\left\lfloor{#1}\right\rfloor}
\newcommand{\step}[1]{\vspace{0.5em}\noindent{#1}}
\newcommand{\quat}[1]{\ensuremath{\mathring{\mathbf{#1}}}}
\newcommand{\norm}[1]{\left\lVert#1\right\rVert}
\newcommand{\ignore}[1]{}
\newcommand{\specialcell}[2][c]{\begin{tabular}[#1]{@{}c@{}}#2\end{tabular}}
\newcommand*\Let[2]{\State #1 $\gets$ #2}
\newcommand{\algorithmicbreak}{\textbf{break}}
\newcommand{\Break}{\State \algorithmicbreak}
\newcommand{\ra}[1]{\renewcommand{\arraystretch}{#1}}

\renewcommand{\vec}[1]{\mathbf{#1}} 

\algdef{S}[FOR]{ForEach}[1]{\algorithmicforeach\ #1\ \algorithmicdo}
\algnewcommand\algorithmicforeach{\textbf{for each}}
\algrenewcommand\algorithmicrequire{\textbf{Require:}}
\algrenewcommand\algorithmicensure{\textbf{Ensure:}}
\algnewcommand\algorithmicinput{\textbf{Input:}}
\algnewcommand\INPUT{\item[\algorithmicinput]}
\algnewcommand\algorithmicoutput{\textbf{Output:}}
\algnewcommand\OUTPUT{\item[\algorithmicoutput]}

\maketitle
\thispagestyle{empty}
\pagestyle{empty}

\begin{abstract}
Forestry machines operated in forest production environments face challenges when performing manipulation tasks, especially regarding the complicated dynamics of underactuated crane systems and the heavy weight of logs to be grasped. 
This study investigates the feasibility of using reinforcement learning for forestry crane manipulators in grasping and lifting heavy wood logs autonomously. We first build a simulator using Mujoco physics engine to create realistic scenarios, including modeling a forestry crane with $8$ degrees of freedom from CAD data and wood logs of different sizes. 
We further implement a velocity controller for autonomous log grasping with deep reinforcement learning using a curriculum strategy. Utilizing our new simulator, the proposed control strategy exhibits a success rate of 96\% when grasping logs of different diameters and under random initial configurations of the forestry crane. 
In addition, reward functions and reinforcement learning baselines are implemented to provide an open-source benchmark for the community in large-scale manipulation tasks. A video with several demonstrations can be seen at \href{https://www.acin.tuwien.ac.at/en/d18a/}{https://www.acin.tuwien.ac.at/en/d18a/}. 
\end{abstract}


\section{INTRODUCTION} \label{Sec:Intro}
Forestry machines are equipped with redundant hydraulic manipulators which are used for manipulation tasks such as grasping and arranging heavy wood logs. 
Although automation has become prevalent across different industries, forestry machines rely primarily on manual operation. 
Operating these machines requires highly skilled operators trained for many hours. 
This is because these machines comprise un-actuated joints and hydraulic actuators with
several constraints, e.g., joint limits, actuator limits, and total pump flow limits. 
Currently, the world is facing a shortage of highly skilled machine operators due to the aging population in Europe and the USA, the high workload, and the extensive training costs. 
As a result, there is an urgent need to develop concepts for the autonomous or semi-autonomous operation of such machines, see 
\cite{ortiz2014increasing}, \cite{kalmari2017coordinated}. 

In recent years, different simulation environments and benchmarking frameworks have been developed in robotics research, e.g.,  \cite{ibarz2021train,sunderhauf2018limits,nguyen2024language,kumar2024robohive,mittal2023orbit,vuong2023grasp}, driven by the emergence of data-driven approaches like machine learning and deep reinforcement learning that heavily rely on training data. This makes the use of simulated environments for training-specific practical tasks popular because given the slow operating speeds of robots and safety considerations, acquiring a large amount of real-world robotic data is a big challenge. However, building a realistic simulator requires the capability to represent physics, dynamics, and sensors within the simulation environment. Despite the inherent challenges, the integration of machine learning, particularly reinforcement learning (RL), with robotics has enabled significant progress in intelligent systems, especially in robotic manipulation~\cite{wang2023dexterous,zhou2023hacman,vuong2024language,vu2025online}. While most RL-based approaches focus on training specific skills for conventional manipulation tasks with popular collaborative robots \cite{mu2021maniskill}, large-scale manipulators, such as those used in forestry, have only recently started to gain attention \cite{andersson2021reinforcement,ayoub2023grasp}. In this work, we prioritize two core objectives: \textit{i)} building a robust, open-source simulation environment tailored for forestry crane operations and \textit{ii)} benchmarking reinforcement learning algorithms on a challenging wood-log grasping task. Our framework not only allows for the realistic simulation of large-scale manipulator tasks but also provides a platform to evaluate and improve RL performance in heavy-duty environments.

\begin{figure*}
\centering
\label{fig:rotbot1}
\scalebox{0.98}{
\def\svgwidth{2\columnwidth}
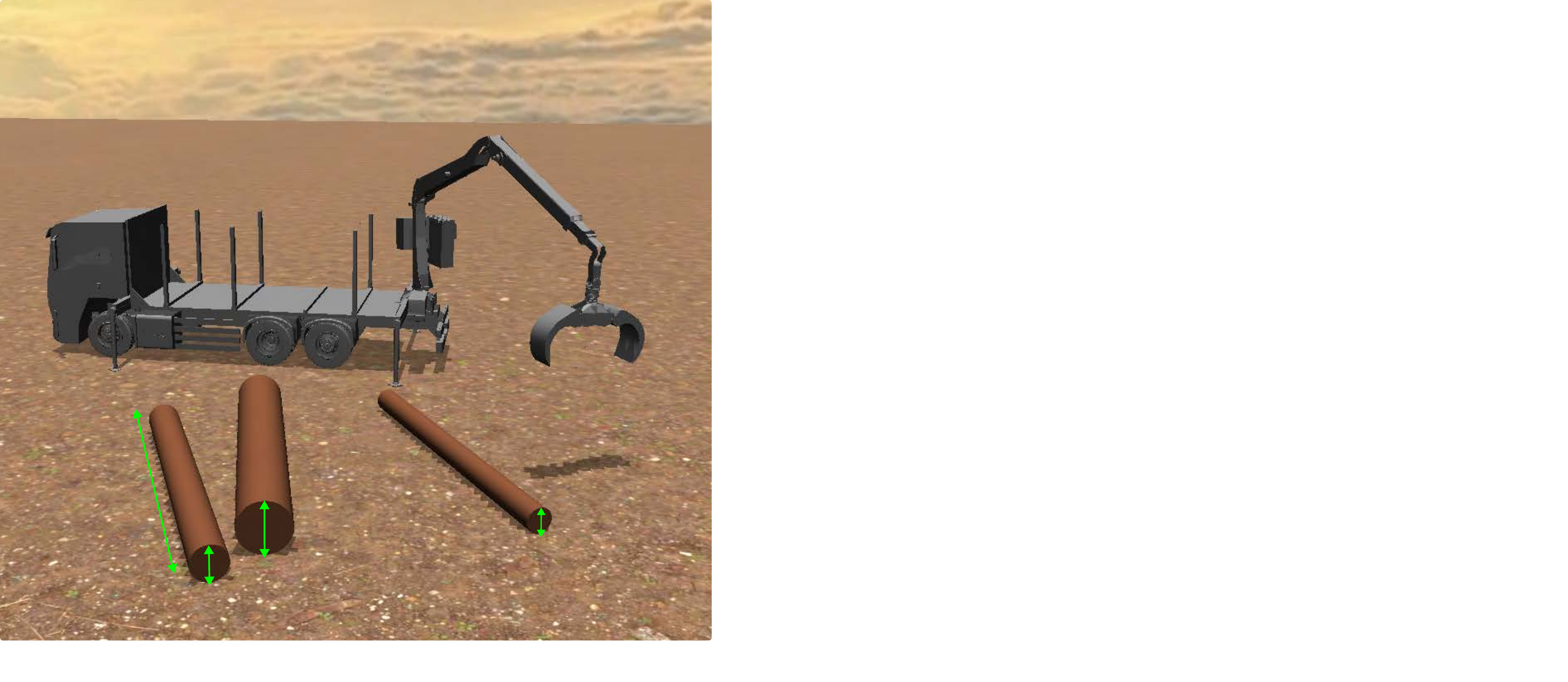
}
\vspace{2ex}
\caption{(a) Snapshot of the simulated environment. While the logs' length is fixed to $l=\SI{2.75}{\meter}$, their diameters are varied in the range $d=[0.3,\:0.8]$ \SI{}{\meter}, (b) Real crane setup in outdoor environment.}
\label{fig: example mujoco}
\vspace{-0.2ex}
\end{figure*}

We consider a popular scenario in the wood-log grasping task: The forestry crane is randomly placed near a wood log location. Our goal is to estimate the wood log's six degrees of freedom (DoF) pose, including its location in the 3D Cartesian space and its orientation, and then develop an RL-based controller to grasp the wood log at its center and lift it. 
An example of the simulation environment is illustrated in Figure~\ref{fig: example mujoco}. 
To facilitate the training, we create a closely matching simulated model and environment based on our real forestry crane setup. The system dynamics and kinematics are verified through a calibration process, which helps to reduce the sim-to-real gap in future applications. 
In the real setup shown in Figure~\ref{fig: example mujoco} (b), the position of the selected tree log is localized using a precise sensor fusion vision system, such as lidar and stereo cameras. Then, the proposed RL method is used to grasp the log. In this way, thanks to its modularity, we can ensure the safety and flexibility of the proposed framework, which are very important in a large-scale environment.

The contributions of our work are as follows:
\begin{itemize}
    \item We present a simulation environment of the large-scale forestry crane that is built from the CAD model of a real forestry crane. 
    \item We propose a new RL algorithm, modified Proximal Policy Optimization (mPPO) by replacing the Gaussian distribution with a restricted distribution and adding a heuristic factor to the trained network proves efficient for the heavy wood-log grasping task. 
\end{itemize}

\section{Related Work} \label{Sec:rw}

In \cite{ortiz2014increasing}, path planning approaches for the crane grapple are presented for moving from the initial configuration to the grasping position.  
Recently, Andersson et al. \cite{andersson2021reinforcement} presented an RL-based solution for the half-loading cycle, i.e. moving the crane to the log and grabbing the log. To simplify the setup, the authors in~\cite{andersson2021reinforcement} only consider a small log next to a forwarder on the ground and the poses of the log are random, however, the size of the log is constant. 

Recently, model-based approaches for controlling a forestry crane are investigated in  \cite{kalmari2014nonlinear} and \cite{hera2015model}, where model-predictive controllers are utilized to reduce the swinging motion of the grapples. A closed-loop RL-based controller has been proposed in \cite{dhakate2022autonomous} for a redundant hydraulic forestry crane for position-tracking tasks. In \cite{andersson2021reinforcement}, the reinforcement learning control for grasping a log is presented, which is most closely related to the present work. The authors utilized Proximal Policy Optimization (PPO) \cite{schulman2017proximal} to train multiple grasping strategies for a forestry crane on AGX dynamics \cite{algoryx}, which is a commercial software for simulating the dynamics of multi-contact systems. 
Although the RL policy in \cite{andersson2021reinforcement} is only trained for a fixed type of wood log, the paper gives good insights into the possibility of employing RL methods for large-scale robots. More recently, Ayoub et al. \cite{ayoub2023grasp} presented a robust grasp planning pipeline, including wood log detection, trajectory planning, and a controller for grasping multiple logs. The work in \cite{ayoub2023grasp} mainly focuses on using a convolutional neural network (CNN) to predict the grasp location and orientation.  
\section{Forestry Crane Simulator} \label{Sec:method}

\subsection{Kinematics}

\begin{figure}
    \centering
    \scalebox{0.8}{
    \includegraphics[trim=5cm 3.5cm 0cm 2cm,clip,scale=0.44]{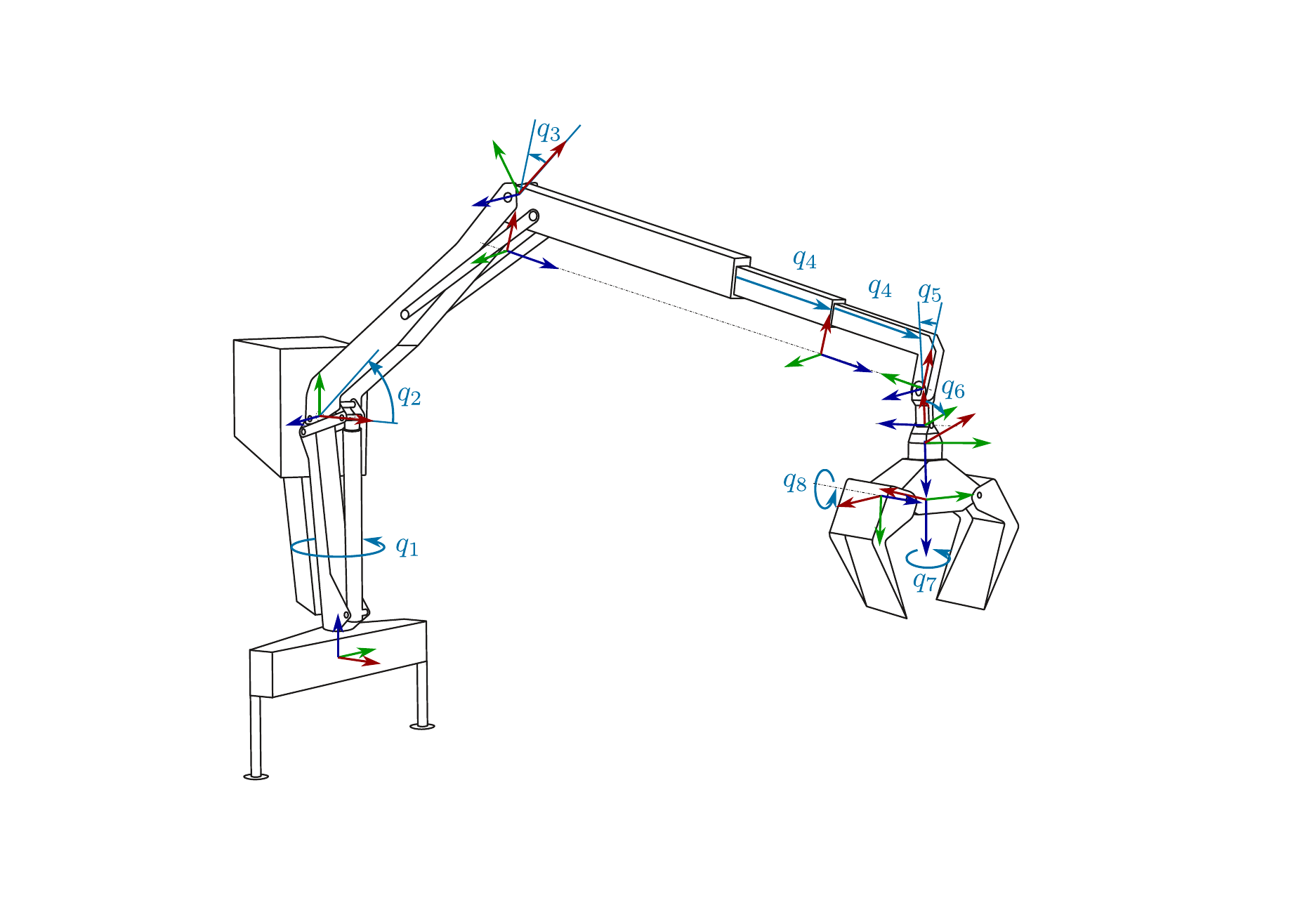}
    }
    \caption{Schematic of the forestry crane \cite{ecker2022iterative}.}
    \label{fig:crane_scematics}
\end{figure}

Figure \ref{fig:crane_scematics} illustrates the schematic of the forestry crane. 
It has eight degrees of freedom (DoFs) $\mathbf{q}^\mathrm{T} = [\mathbf{q}_A^\mathrm{T},\mathbf{q}_U^\mathrm{T}] $ consisting of six actuated DoFs $\mathbf{q}_A^\mathrm{T} = [q_1,q_2,q_3,q_4,q_7,q_8]$ and two unactuated joints $\mathbf{q}_U^\mathrm{T} = [{q}_5,{q}_6]$. Note that there are two pairs of synchronized joints, i.e., the prismatic joint $q_4$ and the revolute joint $q_8$. \corr{In each pair of synchronized joints, the same input is applied to the corresponding actuators; for example, the joint angle $q_8$ at the left- and right-jaw of the grapple in Figure \ref{fig:crane_scematics}. }

The kinematics of the forestry crane are described by transformations from a coordinate Frame $\mathcal{F}_i$ attached to joint $i$ to a coordinate frame $\mathcal{F}_{i-1}$ attached to joint $i-1$
\begin{align}
	\vec{H}^i_{i-1} = \begin{bmatrix}
		\vec{R}_{i-1}^i & \vec{d}_{i-1}^i\\
		\transpose{\vec{0}} & 1
	\end{bmatrix}\in\mathcal{SE}(3) \Comma
\end{align}
where $\vec{R}_{i-1}^i\in\mathcal{SO}(3)$ and $\vec{d}_{i-1}^i\in\mathbb{R}^3$ are the three-dimensional rotation matrix and the three-dimensional translation vector, respectively. 
The coordinate frames are illustrated in Figure~\ref{fig:crane_scematics} according to the \textit{Denavit-Hartenberg (DH) convention} \cite{spong:2006}. 

\subsection{Simulator}
\label{sec: b simulator}

The system dynamics and contacts with the environment are simulated using the open-source MuJoCo \cite{todorov2012mujoco} physics engine. 
An example of the simulated environment is illustrated in Figure \ref{fig: example mujoco}. 
The assembled model of the forestry crane (including the truck) consists of $38$ rigid bodies and $10$ active joints\corr{, including two pairs of synchronized joints}. The total operating weight of the forestry crane is approximately \SI{1981}{\kilo\gram}. 
On standard forestry cranes, hydraulic actuators powered by a pump that is driven by a combustion engine drive the slewing ($q_1$), boom ($q_2$), arm ($q_3$), and prismatic ($q_4$) joints, respectively. 
In order to simplify the model for training purposes, the hydraulic actuators are not explicitly modeled. 
\corr{We assume an (ideal) underlying velocity controller, with the reference velocity for the prismatic joint $q_4$ and the reference rotational velocity for the other joints as inputs. }
Thus, in the simulation environment, the grasping controller for the modeled forestry crane is a fine-tuned PID controller with reference velocities for the actuated joints $\mathbf{q}_A$.  

The wood log position is randomized in a reachable region of the forestry crane, depicted as the yellow region in Fig. \ref{fig: example mujoco}. 
The center of this region is approximately \SI{6.5}{\meter} from the crane's base. 
Additionally, logs are modeled as cylinders with a length of $\SI{2.75}{\meter}$, and the log's diameter varies in the range of $[0.3,0.8] \SI{}{\meter}$. 
In order to prevent overfitting during the training process, the slew angle of the crane is varied in the range $[-2\pi/3, -\pi/3] \SI{}{\radian}$. 
The 6-dimensional contact forces between the grapple and the wood log are computed using the signed distance field (SDF) collision primitive \cite{reiner2011interactive}. 
This is particularly important to maintain the robustness of the simulation since the inner and outer jaw of the grapple have curvy shapes.

\section{Wood-Log Grasping Method}
\label{sec: method}
Since a varied-diameter wood log is considered for the grasping task, the latent Markov decision process (L-MDP) \cite{chen2021understanding,vuong2019pick} is utilized in this work. The reward function and the policy gradient method are briefly discussed in this section.

\subsection{Latent-MDP for forestry crane with varied-diameter wood logs}
The control problem for the grasping task can be modeled as a Markov decision process (MDP), which emulates the interactive learning between the agent (the grasping controller) and the simulated environment. An MDP consists of the $4$-element tuple $(\mathcal{S},\mathcal{A},\mathbf{P},R)$ referring to the state space $\mathcal{S}$, the action space $\mathcal{A}$, the transition probability density function $\mathbf{P}$, and the reward function $R$, respectively. 
At a given time step $t$, the agent, in the state $\mathbf{s}_t \in \mathcal{S}$, selects an action $\mathbf{a}_t \in \mathcal{A}$ to transition to the next state $\mathbf{s}_{t+1}$ 
with the transition probability $\mathbf{P}(\mathbf{s}_{t+1}|\mathbf{s}_t,\mathbf{a}_t)$. 
This results in the immediate reward $R_t$. Note that $\mathbf{P}(\mathbf{s}_{t+1}|\mathbf{s}_t,\mathbf{a}_t)$ is obtained from the simulator introduced in Section \ref{sec: b simulator}. It is worth noting that the transition function $\mathbf{P}$ is uncertain since the contact dynamics are not the same for different wood log dimensions.  
Details on the state $\mathbf{s}_t$, the action $\mathbf{a}_t$, and the reward function $R_t$ for the grasping task with the forestry crane are introduced in the following subsection. 

Since the diameter $d$ of a wood log and its mass vary during the training process, the latent MDP (L-MDP) is utilized, where the log's size can be considered a latent variable. Additionally, individual training episodes are considered as single MDPs with finite length $H$. We denote the L-MDP as $\{\mathcal{L}, p(d)\}$, where $\mathcal{L}$ is the set of single MDPs with different diameters $d$, and $p(d)$ is the \corr{uniform} distribution of the diameter $d$ over $\mathcal{L}$. To this end, the objective of the grasping task is as follows
\begin{equation}
    J(\bm{\pi_\theta}) = \mathrm{E}_{d\sim p(d)}\bigg[\mathrm{E}_{\mathbf{a}_t\sim \bm{\pi_\theta}(.|\mathbf{s}_t)} \Bigg[\sum_{t=0}^{H}\gamma R_t\Bigg] \bigg]\:,
    \label{eq: RL objective}
\end{equation}
\corr{where $\mathrm{E}$ is the expectation function, the symbol ``$\sim$'' denotes the sampling process from the corresponding distribution $\bm{\pi_\theta}$ over the action space $\mathcal{A}$, and $0<\gamma<1$ is the discount factor. The normal distribution is typically used to model $\bm{\pi_\theta}$.} 
Additionally, $\bm{\theta}$ combines the policy parameters that can be weights and biases of a neural network. 
An RL method is utilized to find the optimal policy $\bm{\pi}^*$ that maximizes (\ref{eq: RL objective}) 
\begin{equation}
    \bm{\pi_\theta}^* = \argmax_{\bm{\pi_\theta}} J(\bm{\pi_\theta}) \:.
    \label{eq: RL policy}
\end{equation}
To find the optimal policy (\ref{eq: RL policy}), we utilize the modified version of Proximal Policy Optimization (PPO) as discussed in Subsection \ref{sec: RPPO}. The following subsection presents the details of the observation space, action space, and reward function. 
\subsection{Learning environment for the forestry crane}
\subsubsection{Observations and actions} 
\label{sec: observation}
\begin{figure}[t]
\centering
\scalebox{0.6}{
\def\svgwidth{1\columnwidth}
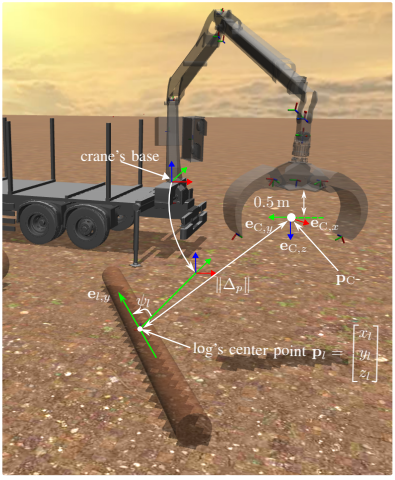
}
\vspace{0.1cm}
\caption{Details of variables used for constructing the observations and reward function.}
\label{fig: rl explained}
\vspace{-0.2ex}
\end{figure}

The observations consist of joint angles of the forestry crane and the actuated joint velocities, i.e., $\mathcal{O} = \{\mathbf{q},\dot{\mathbf{q}}_{A}\}$.  
From the 6 DoF poses of the log's pose w.r.t the crane's base obtained by other algorithms \cite{wen2023bundlesdf,vuong2023grasp}, we consider the reduced poses of 4 DoF $\mathbf{q}_l = [x_{l},y_{l}, z_{l}, \psi_l]^\mathrm{T}$, consisting the 3D Cartesian position of the log's center point $\mathbf{p}_l = [x_{l},y_{l}, z_{l}]^\mathrm{T}$ and the yaw angle $\psi_l$, see Figure \ref{fig: rl explained}. The augmented relative Cartesian distance is computed as
\begin{equation}
    \bm{\Delta}_p =  [x_{l},y_{l}, z_{l} - (d_{max}-z_l)/2]^\mathrm{T} - \mathbf{p}_\mathrm{C}(\mathbf{q}) \:,
    \label{eq: relative distance}
\end{equation}
where $\mathbf{p}_\mathrm{C}(\mathbf{q}) = [p_{\mathrm{C},x},p_{\mathrm{C},y},p_{\mathrm{C},z}]^\mathrm{T}$ results from the forward kinematics
\begin{equation}
    \mathbf{H}_\mathrm{C}(\mathbf{q}) = 
    \begin{bmatrix}
        \mathbf{e}_{\mathrm{C},x} & \mathbf{e}_{\mathrm{C},y} & \mathbf{e}_{\mathrm{C},z} & \mathbf{p}_\mathrm{C} \\
        0 & 0 & 0 & 1
    \end{bmatrix}
\end{equation}
and is located at the center of the grapple, see Figure \ref{fig: rl explained}. Note that $\mathbf{e}_{\mathrm{C},x}$, $\mathbf{e}_{\mathrm{C},y}$, and $\mathbf{e}_{\mathrm{C},z}$ are column vectors of the orientation of the grapple's center. The term $d_{off} = (d_{max}-z_l)/2$ represents an offset in $z$-direction for different log sizes where $d_{max} = 0.8$ is the maximum diameter of the wood log. 
Since $\psi_l$ is the yaw rotation around the $z$-axis of the crane base, illustrated in Fig. \ref{fig: rl explained}, the unit vector $\mathbf{e}_{l,y}$ along the length of the log is computed as
\begin{equation}
    \mathbf{e}_{l,y} = [-\sin(\psi_l), \cos(\psi_l), 0]^\mathrm{T}
\end{equation}
In order to successfully grasp the wood log, the orientation of the grapple must be well-aligned with the wood log, as defined in the following condition
\begin{equation}
     \mathrm{mod}\bigg[\widehat{(\mathbf{e}_{l,y},\mathbf{e}_{\mathrm{C},x})},\pi \bigg] \approx 0 \:, 
     \label{eq: orientation condition}
\end{equation}
where the $\widehat{(\mathbf{e}_{l,y},\mathbf{e}_{\mathrm{C},x})}$ presents the angle between the two vectors $\mathbf{e}_{l,y}$ and $\mathbf{e}_{\mathrm{C},x}$. 
The condition (\ref{eq: orientation condition}) can be normalized as the angle distance function in the form
\begin{equation}
    \Delta_{\psi} = 1- |\mathbf{e}_{\mathrm{C},x}\cdot \mathbf{e}_{l,y} |\:\:,
    \label{eq: angle distance}
\end{equation}
where the symbol ``$\cdot$'' denotes the dot product between two vectors. 
To this end, the observation space also includes the relative distance (\ref{eq: relative distance}) and the angle distance (\ref{eq: angle distance}), i.e., $\mathcal{O} = \{\mathbf{q},\dot{\mathbf{q}}_{A},\bm{\Delta}_p,\Delta_\psi\}$. As ideal underlying velocity controllers are assumed, the action space consists of desired actuated joint velocities $\mathcal{A} = \{\dot{\mathbf{q}}_{A,d}\}$. 

\subsubsection{Reward function}    
The reward function $R$ is designed to gradually guide the grapple along the actions, e.g., approaching, grasping, lifting, and balancing, to achieve the final goal. 
First, the forestry crane can grasp the wood log when the combined weighted distance 
\begin{equation}
    d_\mathrm{combine} = \norm{\bm{\Delta}_p} + \omega_1 \Delta_\psi
    \label{eq: d combine}
\end{equation}
is small enough. Consequently, the associated reward function term reads as
\begin{equation}
    r_{\mathrm{distance}}  = \mathrm{exp}(-\omega_2 d_\mathrm{combine}) \:,
    \label{eq: r_distance}
\end{equation}
where $\omega_1 > 0 $ and $\omega_2 > 0$ are user-defined parameters. 
When the crane approaches the target, the RL agent is encouraged to close the grapple to hold the wood log. The reward function term for this behavior is
\begin{equation}
    r_{\mathrm{grapple}}  = r_{\mathrm{distance}}({q_{8}}/{\overline{q_8}}) + (1-{q_{8}}/{\overline{q_8}})(1-r_{\mathrm{distance}})\:,
\end{equation}
with $\overline{q_8} = \SI{3}{\radian}$ as the limit of the joint angle $q_8$. 
After holding the wood log inside the grapple, the forestry crane proceeds with the lifting action, represented by the reward function term
\begin{equation}
    r_{\mathrm{lift}} = (1- \mathrm{tanh}(\omega_3|z_l-z_{l,d}|))(1-r_\mathrm{grapple})\:\:,
\end{equation}
where $z_{l,d}$ is the desired height of the log and $\omega_3 >0$ is a user-defined parameter. Finally, we encourage the forestry crane to stabilize after grasping the log by using
\begin{equation}
    r_{\mathrm{balance}} = (1 - \mathrm{tanh}(\norm{\dot{\mathbf{q}}_{A,d}}))(1-r_\mathrm{lift}) \:\:.
\end{equation}
Combining all parts, the overall reward function takes the form
\begin{equation}
    R = r_{\mathrm{distance}} + r_{\mathrm{grapple}} + r_{\mathrm{lift}} + r_{\mathrm{balance}} \:\:.
\end{equation}

\subsubsection{Episode termination}
\label{sec: early termination}
Each training episode has a time limitation of $t_{max} = \SI{9}{\second}$. 
Additionally, other termination criteria are listed in the following: 
\begin{itemize}
    \item If the grapple point $\mathbf{p}_C$ is not close to the log's center point $\mathbf{p}_l$,  i.e., $d_\mathrm{combine} < \epsilon$, within $t_{limit} = \SI{6}{\second}$, the episode is early terminated. 
    \item One of the joint limits is violated. 
    \item The log is located more than \SI{8}{\meter} away from the grapple. 
    \item The velocity of the actuated joints exceeds the physical limits, i.e., $|\dot{\mathbf{q}}_A| > \dot{\mathbf{q}}_{A,max}$. 
\end{itemize}
\subsection{Modified proximal policy optimization (mPPO) utilizing Beta distribution}
\label{sec: RPPO}

\begin{figure}[t]
\centering
\scalebox{0.8}{

\def\svgwidth{1\columnwidth}
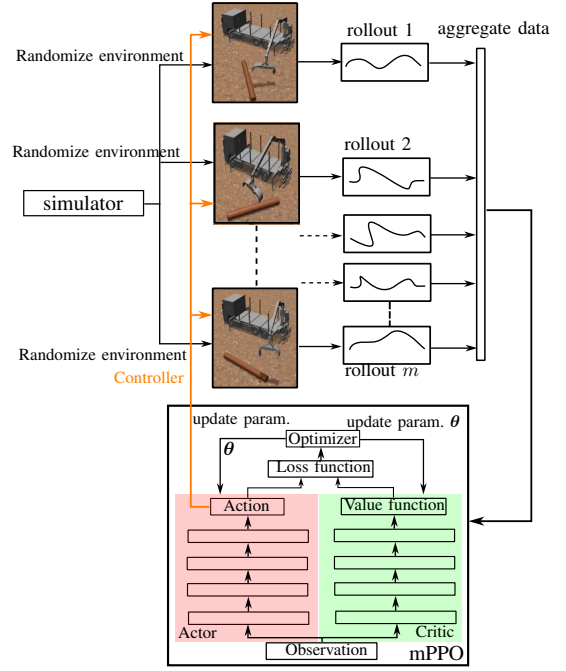
}
\vspace{1ex}
\caption{Overview of the learning process. $m$ randomized environments with different wood log sizes and poses are generated by our crane simulator, presented in Subsection \ref{sec: b simulator}. }
\label{fig: overview learning}
\end{figure}
The overview of the learning process is illustrated in Figure \ref{fig: overview learning}. Using the crane simulator in Mujoco, $m$ parallel environments are sampled to generate rollouts (trajectories) for training the agent. A standard architecture of an actor-critic network used in the PPO algorithm is illustrated on the right-hand side of Figure \ref{fig: overview learning}. Since the details on PPO are omitted, readers are referred to \cite{schulman2017proximal}. Only modifications of the PPO, named mPPO, are presented below. 
    
In deep RL, the policy $\bm{\pi}$ is a neural network with the parameter vector $\bm{\theta}$ that takes the state $\mathbf{s}_t$ as input and outputs the distribution of actions $\bm{\pi_{\theta}}$ modeled as Gaussian distribution in the form
\begin{equation}
    \bm{\pi_{\theta}}(\mathbf{a}_t|\mathbf{s}_t) = \dfrac{1}{\sqrt{2\pi}\bm{\sigma_\theta}}\mathrm{exp}
    \Bigg(
    \dfrac{-(\mathbf{a}_t-\bm{\mu_\theta})^2}{2\bm{\sigma_\theta}}
    \Bigg)\:.
\end{equation}
The control actions $\mathbf{a}_t$ can be sampled in the backpropagation process or the inference process as follows
\begin{equation}
    \mathbf{a}_t = \bm{\mu_\theta}(\mathbf{s}_t) + \bm{\sigma_\theta}(\mathbf{s}_t)\mathcal{N}(0,1)   \:,
\end{equation}
\corr{
where $\bm{\mu_\theta}$ and $\bm{\sigma_\theta}$ are the mean and standard deviation of the Gaussian distribution. Since the control input in $\mathbf{a}_t$ is modeled as a separate distribution, an element-wise product is used for all equations in this subsection.}

However, the control actions of the considered forestry crane are constrained in an admissible range $\underline{\mathbf{a}_t} \leq \mathbf{a}_t \leq \overline{\mathbf{a}_t}$ for safety reasons. Thus, in the mPPO algorithm, the Beta distribution is employed with the probability density function (PDF) \cite{chou2017improving}
\begin{equation}
    \bm{\pi_\theta}(\mathbf{a}_t|\mathbf{s}_t) = \bm{\mathrm{Beta}}(\mathbf{a}_{t,n};\bm{\alpha_\theta},\bm{\beta_\theta})\:\:, \bm{\alpha_\theta}>1, \:\bm{\beta_\theta} >1 \:\:,
\end{equation}
where $\mathbf{a}_{t,n} = \dfrac{\mathbf{a}_t- \overline{\mathbf{a}_t}}{\overline{\mathbf{a}_t} - \underline{\mathbf{a}_t}}$ is the normalized action and 
\begin{equation}
    \bm{\mathrm{Beta}}(\mathbf{a}_{t,n};\bm{\alpha_\theta},\bm{\beta_\theta}) = \dfrac{\bm{\Gamma}(\bm{\alpha_\theta}+\bm{\beta_\theta})}{\bm{\Gamma}(\bm{\alpha_\theta})\bm{\Gamma}(\bm{\beta_\theta})}\mathbf{a}_{t,n}^{\bm{\alpha_\theta}-1}(1-\mathbf{a}_{t,n})^{\bm{\beta_\theta}-1}\:\:. 
\end{equation}
Note that ${\Gamma}(i) = \int_0^\infty j^{i-1}\mathrm{exp}(-j)\mathrm{d}j$ is the Gamma function \cite{davis1959leonhard}. In this way, the action $\mathbf{a}_t$ is always sampled in the admissible range by using the mean of the Beta distribution 
\begin{equation}
    \bm{\mu}_{t,n} = \dfrac{\bm{\alpha_\theta}(\mathbf{s}_t)}{\bm{\alpha_\theta}(\mathbf{s}_t) + \bm{\beta_\theta}(\mathbf{s}_t)} \:\:\:. 
    \label{eq: sampling}
\end{equation}

In the PPO algorithm, the loss function consists of three parts, i.e., the surrogate loss to constrain the policy update, the error term of the value function, and the entropy term to encourage exploration, see \cite{weng2018policy}. 
In addition, in a conventional PPO algorithm, the agent shows more randomness in its actions, but the surrogate loss can constrain the updating policy during the training process. 
In a complex environment with large search areas, exploration is important for an agent like this forestry crane to complete the task. 
Inspired by Robust Policy Optimization (RPO) \cite{huang2022cleanrl}, at each step of the training process, we perturb the sampling action (\ref{eq: sampling}) by random values in the uniform distribution $\mathbf{g} \sim \mathcal{U(-\epsilon,\epsilon)}$ in the form
\begin{equation}
    \mathbf{a}_{t,n} \leftarrow \mathrm{clip}(\mathbf{a}_{t,n} + \mathbf{g},0,1) \:.
    \label{eq: robust ppo}
\end{equation}
The function $\mathrm{clip}$ limits the value of $\mathbf{a}_{t,n}$ in the range of $[0,1]$. In this work, $\epsilon$ is set to $0.1$. 


\section{Simulation results and Analysis} 
\label{Sec:exp}
The forestry crane simulator is built using Mujoco 3.0 \cite{todorov2012mujoco} in an AMD Threadripper 7980X with 64GB memory of RAM and the GeForce RTX 4090. We use the Conjugate gradient (CG) solver \cite{nazareth2009conjugate} with the implicit Euler integration method with a sampling time of $\SI{5}{\milli\second}$ for solving the multi-contact system dynamics. Since the maximum simulation time for an episode is $\SI{15}{\second}$, the maximum episode length is $H = 15/0.005 = 3000$ steps. 
Using the standard PPO algorithm in the open-source package Stable-baseline3 \cite{raffin2021stable}, the mPPO is implemented by considering the Beta distribution and (\ref{eq: robust ppo}). We utilize $42$ parallel environments in the training process of $5\cdot 10^8$ simulation steps for all tested algorithms. The network architecture, see Figure \ref{fig: overview learning}, has $4$ layers with 256 neurons in each layer. The update interval is $1000$ steps for each rollout.  Additionally, ADAM optimizer is utilized \cite{kingma2014adam} with the learning rate $\lambda = 3\cdot 10^{-4}$ and $30$ mini epochs in each optimization batch. Note that the optimizer will take approximately ${5\cdot 10^8}/(42\times1000) \approx 12000$ update steps by the optimizer. 
\begin{figure}
	\centering
 \scalebox{0.94}{
	\includegraphics[width=0.5\textwidth]{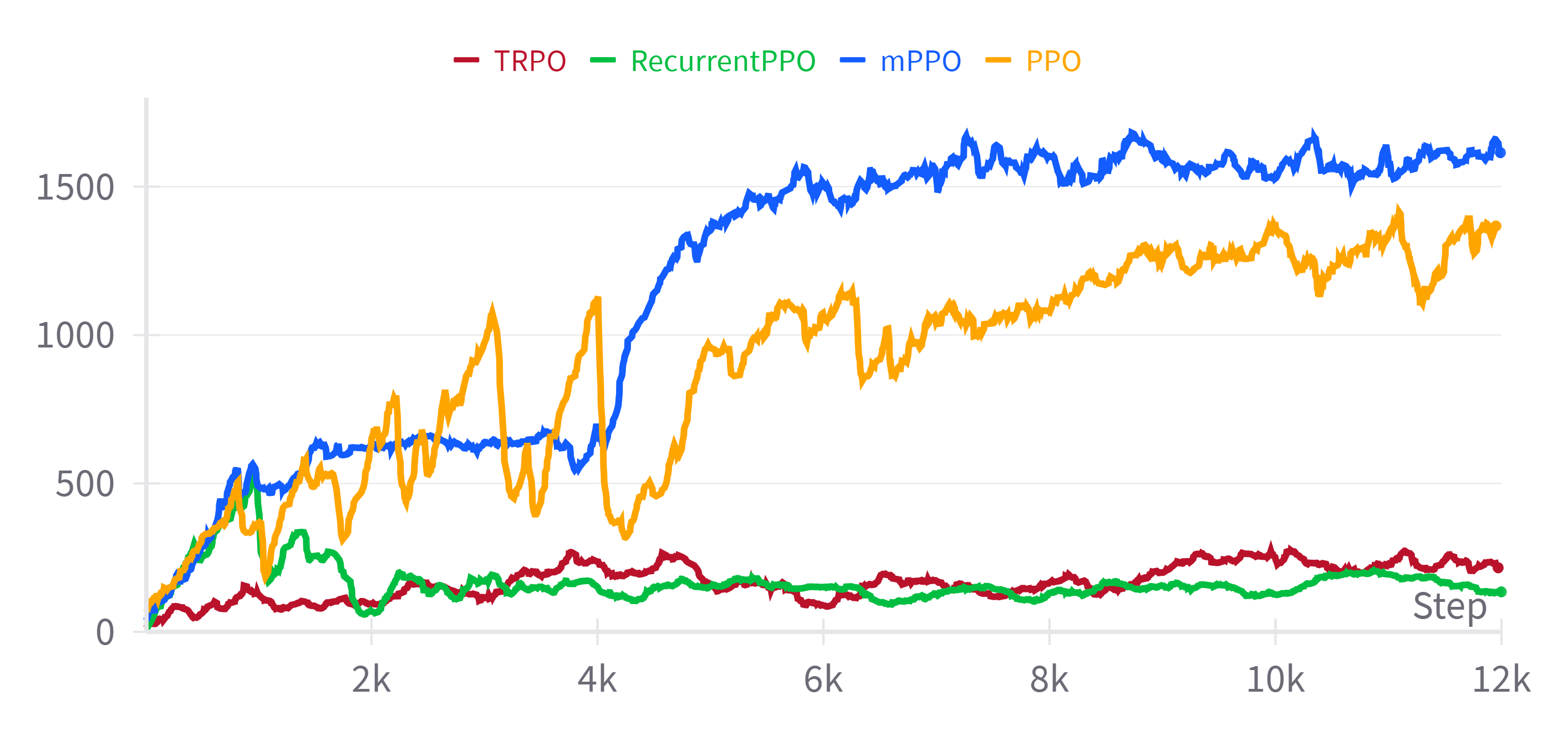}
 }
	\caption{Evolution of cumulative rewards over $12000$ update steps of the optimizer.}
    \label{fig: episode reward}
\end{figure}
\subsection{Benchmarking results}
The cumulative rewards over the whole training process are depicted in Figure \ref{fig: episode reward}, with four different algorithms, i.e., Trust Region Policy Optimization (TRPO) \cite{schulman2015trust}, Recurrent PPO \cite{huang202237}, PPO \cite{schulman2017proximal}, and the proposed mPPO. In the beginning, early terminations, see Subsection \ref{sec: early termination}, are often triggered. This leads to short episode lengths at the beginning of the training process. We achieve better results with the PPO and the mPPO algorithm than the TRPO and the RecurrentPPO algorithms. 
This could be because the agent is not encouraged to explore the large action space by these two latter algorithms. Although we achieve better results with the mPPO than the conventional PPO, the agent sometimes heuristically explores too much, which can lead to a sudden decrease in the average episode length and the cumulative reward. 

It is worth noting that the multi-contact system dynamics are processed in the CPU, while the RL algorithms are executed in the GPU. The computing speeds (including CPU and GPU time) when utilizing mPPO and the PPO algorithms are nearly similar, with $6500$ steps per second. On the other hand, the computing speeds of TRPO and RecurrentPPO are much slower, with $2500$ and $1500$ steps per second. For $5\cdot10^8$ steps, the mPPO and PPO spend approximately $21.5$ hours of training, whereas the TRPO and RecurrentPPO need approximately $56$ and $92.3$ hours, respectively.

\subsection{Insights into the learned grasping skill}
\begin{figure}
\centering
\def\svgwidth{0.8\columnwidth}
\begingroup%
  \makeatletter%
  \providecommand\color[2][]{%
    \errmessage{(Inkscape) Color is used for the text in Inkscape, but the package 'color.sty' is not loaded}%
    \renewcommand\color[2][]{}%
  }%
  \providecommand\transparent[1]{%
    \errmessage{(Inkscape) Transparency is used (non-zero) for the text in Inkscape, but the package 'transparent.sty' is not loaded}%
    \renewcommand\transparent[1]{}%
  }%
  \providecommand\rotatebox[2]{#2}%
  \newcommand*\fsize{\dimexpr\f@size pt\relax}%
  \newcommand*\lineheight[1]{\fontsize{\fsize}{#1\fsize}\selectfont}%
  \ifx\svgwidth\undefined%
    \setlength{\unitlength}{961.39428326bp}%
    \ifx\svgscale\undefined%
      \relax%
    \else%
      \setlength{\unitlength}{\unitlength * \real{\svgscale}}%
    \fi%
  \else%
    \setlength{\unitlength}{\svgwidth}%
  \fi%
  \global\let\svgwidth\undefined%
  \global\let\svgscale\undefined%
  \makeatother%
  \begin{picture}(1,1.15594355)%
    \lineheight{1}%
    \setlength\tabcolsep{0pt}%
    \put(0,0){\includegraphics[width=\unitlength,page=1]{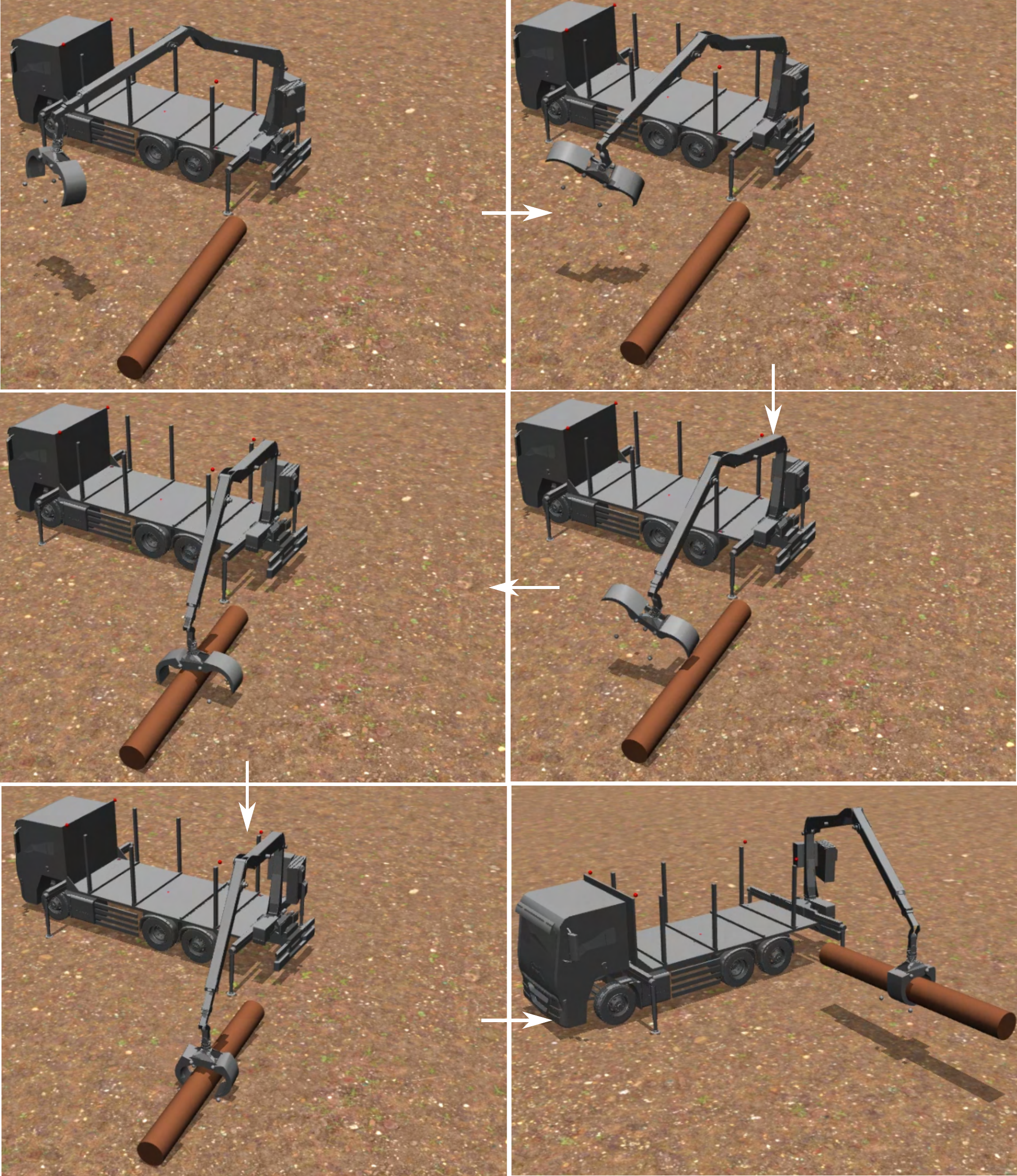}}%
    \put(0.41478964,0.80237049){\color[rgb]{0,0,0}\makebox(0,0)[lt]{\lineheight{1.25}\smash{\begin{tabular}[t]{l}\textcolor{white}{(a)}\end{tabular}}}}%
    \put(0.91610304,0.80301261){\color[rgb]{0,0,0}\makebox(0,0)[lt]{\lineheight{1.25}\smash{\begin{tabular}[t]{l}\textcolor{white}{(b)}\end{tabular}}}}%
    \put(0.91275842,0.40486313){\color[rgb]{0,0,0}\makebox(0,0)[lt]{\lineheight{1.25}\smash{\begin{tabular}[t]{l}\textcolor{white}{(c)}\end{tabular}}}}%
    \put(0.91439481,0.02484381){\color[rgb]{0,0,0}\makebox(0,0)[lt]{\lineheight{1.25}\smash{\begin{tabular}[t]{l}\textcolor{white}{(f)}\end{tabular}}}}%
    \put(0.41236236,0.02523764){\color[rgb]{0,0,0}\makebox(0,0)[lt]{\lineheight{1.25}\smash{\begin{tabular}[t]{l}\textcolor{white}{(e)}\end{tabular}}}}%
    \put(0.40926612,0.41741162){\color[rgb]{0,0,0}\makebox(0,0)[lt]{\lineheight{1.25}\smash{\begin{tabular}[t]{l}\textcolor{white}{(d)}\end{tabular}}}}%
  \end{picture}%
\endgroup%

\vspace{0.2cm}
\caption{Sequence of images from a grasping demonstration with the log diameter $d = \SI{0.4}{\meter}$. }
\label{fig: grasping skill 1}
\end{figure}
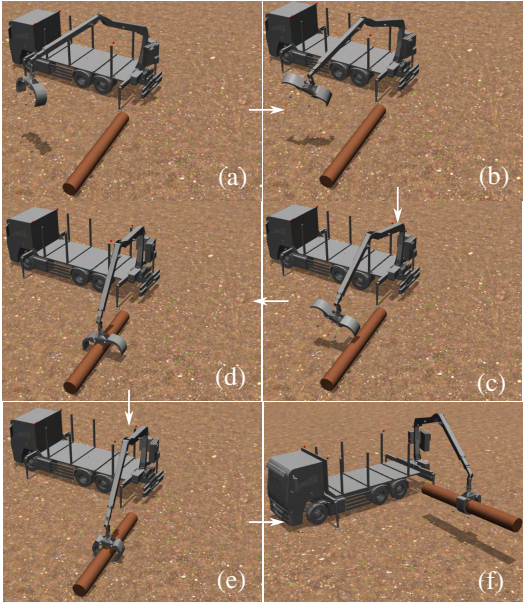

In Figure \ref{fig: grasping skill 1}, an example of a demonstration is performed by the trained agent with mPPO for grasping a wood log with a diameter $d=\SI{0.4}{\meter}$. The forestry crane quickly approaches the log and aligns the grapple with the log's orientation, see Figure \ref{fig: grasping skill 1}(a)-(c). 
It is worth noting that the agent does not use a conventional grasping technique, such as perfectly aligning the grapple with the tree log. Instead, the agent approaches the tree log in a versatile way while moving as long as the vector $\mathbf{e}_{C,x}$ is parallel to the unit length vector of the tree log $\mathbf{e}_{l,y}$, as defined by (\ref{eq: angle distance}). Whenever the combined distance $d_\mathrm{combine}$ from (\ref{eq: d combine}) is below a certain threshold, i.e., $r_\mathrm{distance} \rightarrow 1$, see (\ref{eq: r_distance}), the grapple starts closing, followed by a lifting action; see Figure \ref{fig: grasping skill 1}(d)-(f). 
\definecolor{transfertoclient}{HTML}{ABDDA4}
\definecolor{rendering}{HTML}{2B83BA}

\begin{table}
\caption{Statistical results for grasping logs with different diameter $d$.} 
\vspace{1ex}
\centering
\renewcommand
\tabcolsep{4.5pt}
\hspace{1ex}
\begin{tabular}{@{}rcccccc@{}}
\toprule
$d$ in \SI{}{\meter} & 0.3 & 0.4 & 0.5 & 0.6 & 0.7 & 0.8 \cr 
\midrule
success rate (\%)  & 94 & 96 & 98 & 98 & 96 & 96 \\
\bottomrule
\end{tabular}
\label{table:stats log size}
\end{table}
\subsection{Statistical results}
\label{sec: stats MC}
To challenge the agent, we run a Monte Carlo simulation with six batches with log diameters in the range $d \in \{0.3,...,0.8\} \SI{}{\meter}$. We collect $100$ trials in each batch by randomizing the initial configuration for the forestry crane and the log's pose. 
A trial is considered a success if the agent can reach the log within the time limit of \SI{6}{\second} and fully grasp the log in $\SI{9}{\second}$. Even if the agent can completely capture the tree log within the time limit, if it misses the center of the wood log by a particular threshold value, we also consider this a failed trial. In Table \ref{table:stats log size}, the statistical results of the mentioned Monte Carlo simulation are shown. For the smallest log diameter $d = \SI{0.3}{\meter}$, there are $6$ failed trials in which slowly closing the grapple behavior leads to exceeding the time limit of $\SI{9}{\second}$. With larger logs, failures only occur if the agent misses the center of the log, followed by the excessive swinging motion of the unactuated joints during lifting. Overall, the trained agent achieves a success rate of over $96\%$  for this Monte Carlo simulation. 

\subsection{Bridging Sim-to-real aspect} 
The augmented relative Cartesian distance (\ref{eq: relative distance}) and the angular distance function (\ref{eq: angle distance}) are important as they lead the agent directly to the goal, especially in model-free RL. 
The same assumption is commonly employed in indoor robot manipulation tasks \cite{mu2021maniskill} or large-scale robot manipulation \cite{andersson2021reinforcement}. Learning with solely visual input only works in generative AI scenarios where the agent can learn from expert demonstration while observing multimodal data, e.g., RGB-D and force/torque profiles. 

However, we recognized that assuming a known log's pose, which leads to (\ref{eq: relative distance}) and (\ref{eq: angle distance}), significantly contributes to the sim-to-read gap. 
In order to account for this effect, we add measurement errors to (\ref{eq: relative distance}) and (\ref{eq: angle distance}) in the form
\begin{subequations}
\begin{align}
    \bm{\Delta}_p &\leftarrow  \bm{\Delta}_p + \epsilon_p\bm{\Delta}_p s(\bm{\Delta}_p)\:, \\
    \Delta_\psi &\leftarrow  \Delta_\psi + \epsilon_o\Delta_o s(\bm{\Delta}_p) \:,
\end{align}
\end{subequations}
where $\epsilon_p$ and $\epsilon_o$ are random values drawn from a uniform distribution $U(-0.1,0.1)$. Note that $s(\bm{\Delta}_p)$ is the quadratic decay function, expressed as $s(\bm{\Delta}_p) = (||\bm{\Delta}_p||/{d_{max}})^2$, with $d_{max}=\SI{8}{\meter}$ in our setup. 
This is a practical assumption since, in our real setup, two RGB-D cameras, attached to the control box and the gripper, are employed for log pose detection, see Figure \ref{fig: example mujoco}(b). The same Monte Carlo simulation as in Subsection \ref{sec: stats MC} is performed with six batches (100 trials in each batch) of log diameters $d \in\{0.3,...,0.8\}$. The statistical results of this test are shown in Table \ref{tab:table-new}. Most of the failed cases with small logs, $d= 0.3$ and $d=0.4$, are due to exceeding the time limit of $\SI{9}{\second}$. On the other hand, with larger diameter logs $d \geq 0.5$, failed cases are caused by a miss-aligned grasp due to an orientation error. Overall, the agent achieves a success rate of $\approx 92\%$ for all cases. 
\begin{table}
\caption{Statistical results of adding errors to the pose measurement.} 
\vspace{1ex}
\centering
\renewcommand
\tabcolsep{4.5pt}
\hspace{1ex}
\begin{tabular}{@{}rcccccc@{}}
\toprule
$d$ in \SI{}{\meter} & 0.3 & 0.4 & 0.5 & 0.6 & 0.7 & 0.8 \cr 
\midrule
success rate (\%)  & 88 & 94 & 93 & 93 & 92 & 91 \\
\bottomrule
\label{tab:table-new}
\end{tabular}
\end{table}

\section{Conclusion}\label{Sec:con}
This work introduces a benchmarking for model-free varying-diameter log-grasping with a forestry crane, including the structure of the environment, design of reward functions, and a modified proximal policy optimization (mPPO) algorithm. Under the assumption that the log pose is given, extensive simulations are presented to show the effectiveness of the reward shape and the exploration capability of the mPPO over other algorithms. The overall success rate of the grasping task of varying-diameter wood logs, varying log poses, and randomized initial configurations of the forestry crane exceeds $96\%$. 

\textbf{Limitation.} Although our method shows promising results, we recognize many aspects that require further attention, particularly regarding the sim-to-real gap. For instance, while the simulation offers many benefits, real-world uncertainties such as sensor noise, actuation delays, and unexpected disturbances will require more robust handling. The computational efficiency, especially the training time, can be further optimized by leveraging GPU acceleration. Additionally, incorporating transfer learning techniques may help improve the generalization to physical systems. In future work, we will focus on deploying the learned model in real-world demonstrations and aim to refine the agent’s ability to adapt to dynamic, unpredictable conditions.



\bibliographystyle{class/IEEEtran}
\bibliography{class/reference}
\end{document}